# Continuous wavelet transform of multiview images using wavelets based on voxel patterns

**VLADIMIR SAVELJEV**[*]

*Public Safety Research Center, Konyang University, Nonsan, Chungcheongnam-do 32992, Republic of Korea*
*\*saveljev.vv@gmail.com*

**Abstract:** We propose the multiview wavelets based on voxel patterns of autostereoscopic multiview displays. Direct and inverse continuous wavelet transforms of binary and gray-scale images were performed. The input to the inverse wavelet transform was the array of wavelet coefficients of the direct transform. A restored image reproduces the structure of the multiview image correctly. Also, we modified the dimension of the parallax and the depth of 3D images. The restored and modified images were displayed in 3D using lenticular plates. In each case, the visual 3D picture corresponds to the applied modifications. The results can be applied to the autostereoscopic 3D displays.

## 1. Introduction

We made a transform of the multiview (MV) image in the image plane of an autostereoscopic three-dimensional (3D) display [1]-[3] with a lenticular/barrier plate. At the same time, multiview, integral, plenoptic, and light-field displays have a similar structure of the image. (In plenoptic case, this is called "raw image".) Some peculiar arrangements comprising multiple repeated smaller pieces can be recognized in the image plane of mentioned devices. These arrangements can be treated as combinations of so-called voxel patterns (VP) [4], [5], a sort of elementary particles in the image plane, similar Fresnel patterns in Fresnel holograms [6], [7]. Also important is that the image plane of autostereoscopic 3D display (ASD) consists of the logical cells, one cell per lenticular/barrier element.

In 3D imaging, the wavelets are used for the image fusion [8], gesture recognition [9], video coding [10], [11], etc. However, the known wavelets are mostly used. We propose the specially designed wavelets for MV images of ASD to deal with MV images easier. The wavelets were inspired by the structure of the image in the image plane. VPs are series of rectangular pulses. Previously developed MV wavelets for ASD were based on Haar [12], B-spline [13], and Marr wavelets [14]. The Haar-based MV wavelets are directional, while the MV image does not show directionality. As compared to the MV wavelets based on B-spline and Marr functions, the proposed wavelets have an essentially simple structure, although technically, the PB MV wavelet is not a wavelet packet as [13], [14], rather individual function for each *n*, as in [4], [5].

In this paper, we chose VPs [4], [5] as basic functions for wavelets. A simple structure of the pattern-based (PB) wavelets means potentially fast real-time implementations.

We made the continuous wavelet transform (CWT) by distance planes and obtain the two-dimensional (2D) array of the wavelet coefficients for each plane comprising the 3D array of the coefficients. Then, the 3D array can be either used for the analysis or the inverse wavelet transform directly or alternatively, it can be modified before the inverse transform. The result is another 3D image.

## 2. Background. Patterns and their Fourier spectra

The basic element of VPs is the symmetric rectangular unit pulse of the unit width commonly defined as,

$$\Pi(x) = \begin{cases} 0, |x| > \frac{1}{2} \\ \frac{1}{2}, |x| = \frac{1}{2} \\ 1, |x| < \frac{1}{2} \end{cases} \quad (1)$$

The main properties of VPs [4], [5] in MV images are listed below; also see Fig. 1. We imply the cell size equal to 1, and the intensity range of the image is between zero and one [0, 1]. Let $d$ be a discrete distance from the screen [14]. There are positive and negative distances at opposite sides of the screen of ASD (in front and behind). Each VP corresponds to a discrete depth plane, and the $d$-th pattern consists of $n = |d|$ smaller pieces of the identical amplitude and width, $n = 1, 2, \ldots$, all partitions are non-negative and their summed area is equal to one (the size of the cell).

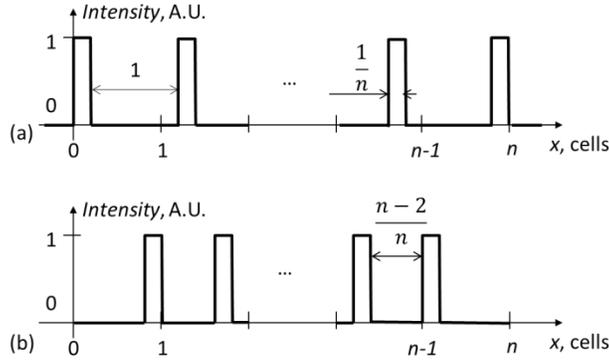

Fig. 1. Voxel patterns of the $d$-th order: (a) $d > 0$, in front of screen, (b) $d < 0$, behind screen.

The width of each pulse is equal to $1/n$-th part of the cell; we will refer these pulses to as partitions of the cell or just partitions. Each $n$ corresponds to two depth planes $+n$ and $-n$ lying at predefined distances [14]. The first depth plane (the sign does not matter) is the screen itself, the zeroth plane is omitted. (Zero partition means an absence of the pattern.)

In VPs for positive and negative planes with the same $n$ (i.e., across the same number of the cells), the partitions are distributed differently; see Fig. 1. Specifically, the first and last partitions are aligned either to the outer or to the inner boundaries of the outmost (leftmost and rightmost) cells of the pattern. There are two cases: (a) a wide gap and the maximum separated pulses correspond to the positive distances in front of the screen, (b) a narrow gap and the maximum close pulses correspond to the negative distances behind the screen. We may call these cases as close and separated pulses, resp.

The single pulse of the width $\tau$ and the amplitude $a$ is expressed as

$$p_0 = a \cdot \Pi\left(\frac{x}{\tau}\right) \quad (2)$$

Note that in this paper, the capital Greek letter $\Pi$ means the rectangular pulse function, but not a product.

VP is a finite series of pulses with the unit height and the shrunk width of $1/n$. Similar to [12], VP can be written as a linear combination of the identical rectangular pulses,

$$P_n(x) = \sum_{i=1}^{n} \Pi(nx - ip_n - s_n), n > 1 \quad (3)$$

where the initial phase of the first pulse (i.e., the position of its center) is

$$s_n = 1 \pm \frac{1}{2n} = \begin{cases} 1 + \frac{1}{2n} \equiv \frac{1}{2n}, d > 0 \\ 1 - \frac{1}{2n}, d < 0 \end{cases} \quad (4)$$

and the period of pulses (the distance between the centers of repeated pulses) is

$$p_n = \frac{n \pm 1}{n} \quad (5)$$

The graph of the period depending on $n$ is shown in Fig. 2.

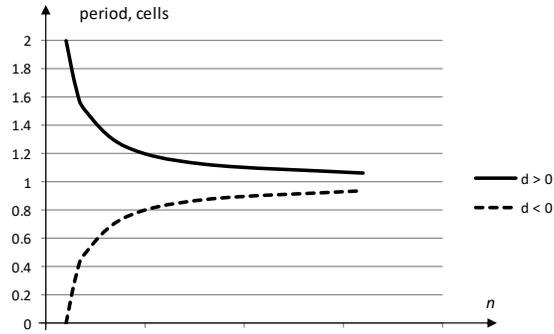

Fig. 2. Period of positive and negative VPs.

The gap between the pulses of VP is

$$\delta_n = 1 - \frac{1}{n} \pm \frac{1}{n} = \begin{cases} 1, d > 0 \\ 1 - \frac{2}{n}, d < 0 \end{cases} \quad (6)$$

Depending on the sign of $d$, the left/right edge of the first/last pulse coincides with the left/right edge of the cell.

Two particular cases of VPs with $n = 1$ and $n = 2$ are shown in Fig. 3.

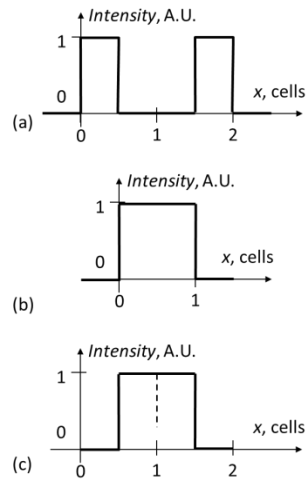

Fig. 3. Particular VPs: (a), (c) second order (two cases: positive in (a) and negative in (c)); (b) first order (one case).

In the case of $n = 1$, the same pattern matches both positive and negative distances $\pm 1$; neither period nor gap have a meaning. The initial phase is,

$$s_1 = \frac{1}{2} \tag{7}$$

Therefore,

$$P_1(x) = \Pi\left(x - \frac{1}{2}\right) \tag{8}$$

VP with $n = 2$ obeys the general rule but may look uncertain, so we explain this case separately. The initial phase and the period of the pulses by Eqs. (4), (5) are,

$$s_2 = \begin{cases} \frac{1}{4}, d > 0 \\ \frac{3}{4}, d < 0 \end{cases} \tag{9}$$

$$p_2 = \begin{cases} \frac{3}{2}, d > 0 \\ \frac{1}{2}, d < 0 \end{cases} \tag{10}$$

Besides, the gap $\delta_{2-}$ at the second negative distance is 0, i.e., two adjacent pulses of the width ½ compose a single pulse of the total width 1, as shown by the dashed line in Fig. 3(b).

Furthermore, let's calculate the Fourier transform (FT) of VPs. Namely, FT of the single pulse Eq. (2) is,

$$F(\omega) = a\tau \cdot sinc\frac{\omega\tau}{2} \tag{11}$$

where the unnormalized sinc function is

$$sinc(x) = \frac{\sin x}{x} \tag{12}$$

Three particular cases of FT (which will be used in this paper later) are as follows.

(a) an elementary pulse with $\tau = 1/n$, $a = A$,

$$F_1 = A\frac{1}{n}sinc\frac{\omega}{2n} = A\frac{1}{n}\frac{\sin\frac{\omega}{2n}}{\frac{\omega}{2n}} = 2\frac{A}{\omega}\sin\frac{\omega}{2n} \tag{13}$$

(b) a typical combination of the elementary pulses (a pair of narrow pulses) with $\tau = 1/n$, $a = A$, symmetrically displaced around the origin by $\pm s$. Recall that FT of a function shifted by $s$ is the FT of the original function multiplied by the exponential,

$$F_s(\omega) = a\tau e^{-i\omega s} sinc\frac{\omega\tau}{2} \tag{14}$$

Thus,

$$F_2 = A\frac{1}{n}sinc\frac{\omega}{2n}(e^{-is\omega} + e^{is\omega}) = 2A\frac{1}{n}sinc\frac{\omega}{2n}\sin(s\omega) \tag{15}$$

$$F_2 = 4\frac{A}{\omega}\sin\frac{\omega}{2n}\sin(s\omega) \tag{16}$$

(c) a bias of a wavelet (a single "shallow" pulse) with $\tau = q$, $a = A/q$,

$$F_- = 2\frac{A}{q\omega}\sin\frac{q\omega}{2} \tag{17}$$

Consequently, the FT of VP of the first order is,

$$F_{p1} = ATsinc\frac{\omega}{2} \tag{18}$$

Then, there are two pulses $\tau = ½$ in VP of the second order, which are displaced either by ¼ or by ¾, as shown in Figs. 3(a), (c). In the case of the adjacent pulses ($s = ¼$), By Eq. (16), FT is,

$$F_{p2-} = 4\frac{A}{\omega}\sin\frac{\omega}{4}\cos\frac{\omega}{4} = 2\frac{A}{\omega}\sin\frac{\omega}{2} = ATsinc\frac{\omega}{2} \tag{19}$$

For graphical illustrations of these cases, see Fig. 4. Note that Eq. (19) is identical to Eq. (18) for VP at $d = 1$. In the case of separate pulses ($s = ¾$), FT of the pattern is,

$$F_{p2+} = 2\frac{A}{\omega} \sin\frac{\omega}{2}\left(4\cos^2\frac{\omega}{4} - 3\right) \qquad (20)$$

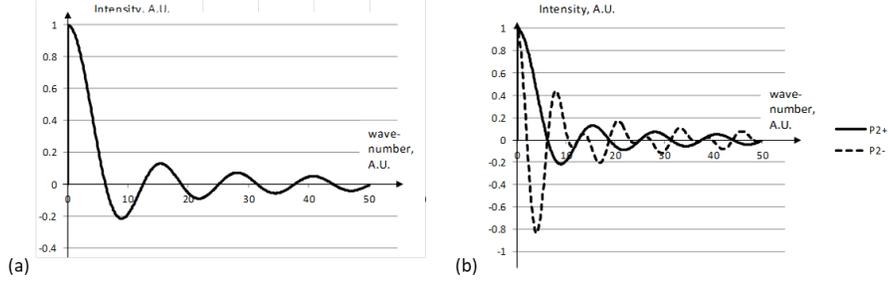

Fig. 4. FT of VPs of (a) first and (b) second orders.

FP of VPs of higher orders can be described similarly, as the sum of pairs Eq. (16), conditionally plus the central pulse Eq. (13) for odd $n$,

$$F_{pn} = \begin{cases} \sum_{i=0}^{\frac{n}{2}} F_2 - F_-, & n \text{ even} \\ \sum_{i=0}^{\frac{n}{2}} F_2 - F_- + F_1, & n \text{ odd} \end{cases} \qquad (21)$$

To illustrate Eq. (21), FT of VPs up to 16th order are shown in Fig. 5.

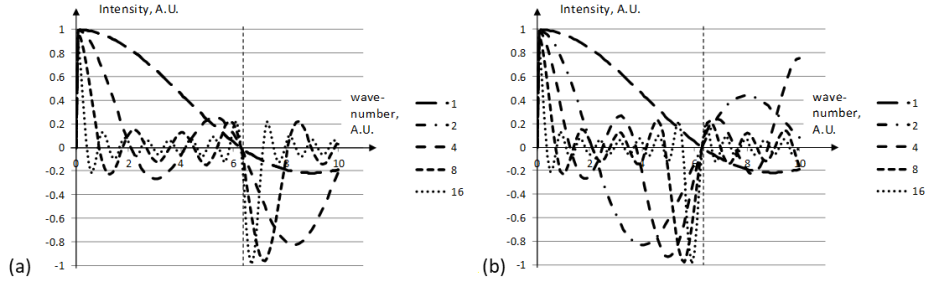

Fig. 5. FT of VPs up to 16th order: (a) negative depth, (b) positive depth.

Note that the minimum (the second extremum after DC) of the FT of the close pulses is on the right side from the dotted vertical line $2\pi$ in Fig. 5, while the second extremum of the separated pulses on the left side. This corresponds to the periods of VPs by Eq. (5) shown in Fig. 2.

## 3. Pattern-based wavelets

Now, let's build the wavelets with the compact support. These wavelets were inspired by the VPs described in the previous Section and the french hat wavelet [15], [16] see Fig. 6. One can alternatively consider this wavelet as a positive rectangular pulse biased ½ down or as two adjacent Haar wavelets of the opposite signs.

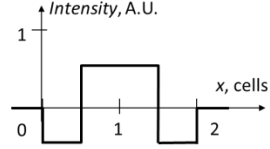

Fig. 6. French hat wavelet.

Generally, the wavelets should satisfy some important properties, namely, the zero mean (no bias), the square norm one (fixed power/energy); the orthogonality is not a requirement for the continuous wavelet transform (CWT).

The french hat wavelet obviously satisfies the requirements. It will be the MV wavelet of the first order. Then, one can make the higher order wavelets to satisfy the zero mean property by subtracting a small constant (bias) from the whole VP across a limited interval (compact support) of $n$ cells (the support of the wavelet). However, after that, the square norm one condition (satisfied in VPs) becomes violated. Nevertheless, by changing the amplitude of the biased VP, this condition can be satisfied again. Then, we have a wavelet with a proper bias and normalization coefficient satisfying the necessary conditions.

Consider two cases: the one-dimensional (1D) wavelets for ASD with the horizontal parallax only (HPO) and the two-dimensional (2D) wavelets for ASD with so-called full parallax (FP). The spectra will be also analyzed.

### 3.1 1D wavelets

HPO wavelets are the functions of one coordinate. As intended, we take VP, bias, and normalize it; thus we obtain the MV wavelet satisfying the conditions of zero mean and square norm one.

According to this suggestion, the general formula for the 1D MV wavelets of the order $n$ is as follows,

$$\psi_{1Dn}(x) = c_{n1D}\big(b_{1Dn} + P_n(x)\big) = c_{n1D}\big(b_{1Dn} + \sum_{i=0}^{n} \Pi(nx - ip_n - s_n)\big) \quad (22)$$

where $c_{1Dn}$ is the normalization coefficient, $b_{1Dn}$ is the bias; the VPs $P_n$ with the period $p_n$ and phase $s_n$, are given by Eqs. (3)-(5). The coefficient and bias will be calculated in this Section.

The overall picture of the PB wavelets built this way is shown in Fig. 7, two particular cases ($n = 1$ and $n = 2$, $p = q = 2$) in Fig. 8.

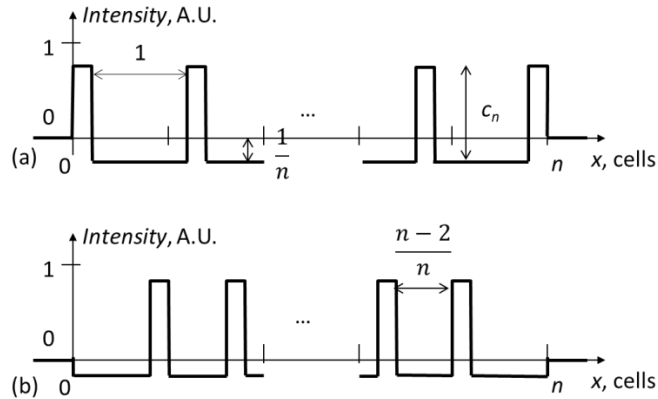

Fig. 7. 1D PB MV wavelets: (a) positive distance, (b) negative distance.

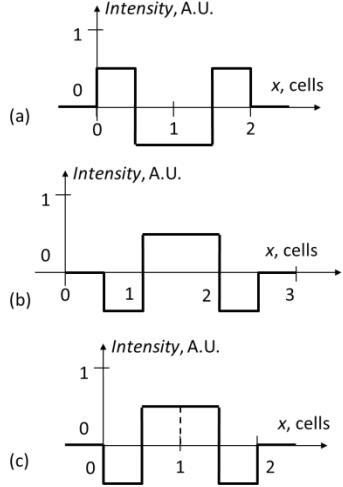

Fig. 8. Particular cases of 1D PB MV wavelets: (a) $n = 2$, (b) $n = 1$, (c) $n = -2$. In all three cases, bias = -½.

Note the different alignment of the pulses to the cells. The wavelet $\psi_1$ shown in Fig. 8(b) is the single rectangular pulse biased by ½; compare it with the French hat Fig. 6 to find a full identity of the shape (the phase is different, though). The wavelets $\psi_2$ are negative copies of each other, $\psi_{2+} = -\psi_{2-}$, see Figs. 8(a), (c). Also note that $\psi_1$ is $\psi_{2-}$ shifted by ½ of the cell (in the former, the positive pulse occupies the whole cell; in the latter, the pulse is shared between two cells, exactly as in the French hat wavelet).

For the 1D PB wavelet of $n$-th order occupying $n$ cells, the bias for both positive and negative distances is,

$$b_{1Dn} = \begin{cases} -\frac{1}{2}, n = 1 \\ -\frac{1}{n}, n > 1 \end{cases} \qquad (23)$$

Particularly, $b_{1D1} = b_{1D2+} = b_{1D2-} = 1/2$.

With the bias by Eq. (23), PB wavelets have zero mean. Let's calculate their normalized power (energy).

In each case, there are $n$ peaks (their total summed width = 1) of the height of $1-1/n$ plus the bias (which height = $1/n$) spread across $n-1$ cells remaining for the bias. Then,

$$\|f\|^2 = \left(\left(1 - \frac{1}{n}\right) \cdot 1 + \frac{n-1}{n}\right)^2 = 4\left(\frac{n-1}{n}\right)^2 \qquad (24)$$

Therefore, the normalization coefficient of 1D case is,

$$c_{1Dn} = \frac{1}{\|f\|} = \begin{cases} 1, n = 1 \\ \frac{1}{2}\frac{n}{n-1}, n > 1 \end{cases} \qquad (25)$$

With $n = 1$, we may use the coefficient for $n = 2$, due to their graphical identity. The normalization coefficient of these three cases = 1: $c_{1D1} = c_{1D2+} = c_{1D2-} = 1$.

Based on the above formulas Eqs. (23), (25), the general formula Eq. (22) for PB wavelet can be rewritten,

$$\psi_{1Dn}(x) = \begin{cases} -\frac{1}{2} + \Pi\left(x - \frac{1}{2}\right), n = 1 \\ \frac{1}{2}\frac{n}{n-1}\left(-\frac{1}{n} + \sum_{i=0}^{n} \Pi\left(nx - i\frac{n+1}{n}\right)\right), n > 1 \end{cases} \qquad (26)$$

Now, let's find FT of the wavelets. For the first order (the single pulse plus the negative bias with $n = 1$, $q = 2$), FT by Eqs. (13), (17) is

$$F_{\psi 1} = 2\frac{A}{\omega}\mathrm{sinc}\frac{\omega T}{2} - \frac{A}{\omega}\mathrm{sinc}(\omega T) = 2\frac{A}{\omega}\sin\frac{\omega T}{2}\left(1 - \cos\frac{\omega T}{2}\right) \quad (27)$$

It is shown in Fig. 9 (a).

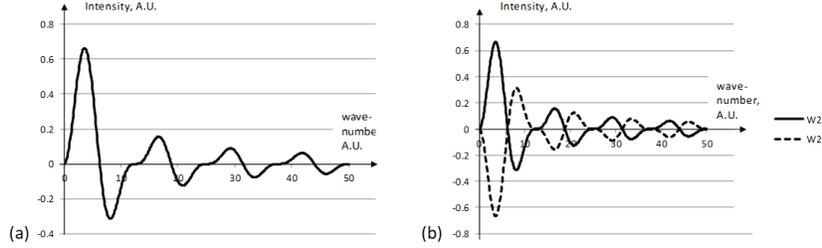

Fig. 9. FT of wavelet of (a) first and (b) second orders.

Note the absence of DC component in FT of the wavelet; except that the graph is quite similar to the FT of VP; cf. Fig. 4(a).

Similar to the VPs of the second order, in the corresponding PB wavelets, two pulses are biased by -½, but are either adjacent or separated. FTs of both cases coincide (except for the sign),

$$F_{\psi 2} = \pm 2AT\,\mathrm{sinc}\frac{\omega T}{2}\sin^2\frac{\omega T}{4} \quad (28)$$

Again, both FTs have 0 at the origin; see Fig. 9(b). Also, the moduli of all three mentioned FT for $n = 1, 2$ are identical.

FTs of the PB wavelets of the higher orders can be found similarly to VPs as sums of FTs of several paired pulses for the even $n$ (plus FT of the centered pulse for the odd $n$) using Eqs. (13), (16), (17). The calculated spectra of the wavelets are shown in Fig. 10.

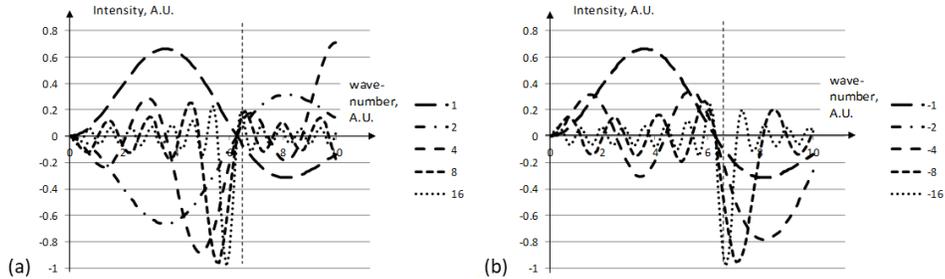

Fig. 10. FT of wavelets up to 16th order: (a) negative $d$, (b) positive $d$ (recall that FT($\psi_1$) = FT($\psi_2$), and $\psi_1 = - \psi_2$).

Note some particular regions in the spectra, where FT is zero or close to zero. The regions in Fig. 10 are as follows: near zero, exactly on the line $2\pi$, and a high-frequency range $> 3\pi$. These spectral regions are not covered by the proposed wavelets, and in general, may provoke negative effects. Thus, we might need some special means to cover these "holes" in the spectrum. Namely, the first region can be covered by the wide single positive pulse of the double width, the second region by a series of the short pulses with period = 1, and the third one by a chirp function which period changes from ½ to the minimal potential period (in a digital screen, 2 pixels), see Fig. 11. The centers of the pulses of the infinite and chirp signals are aligned to the edge of the cell (at least the first pulse of each chirp frequency); the number of pulses of these signals should be enough to cover the whole MV image.

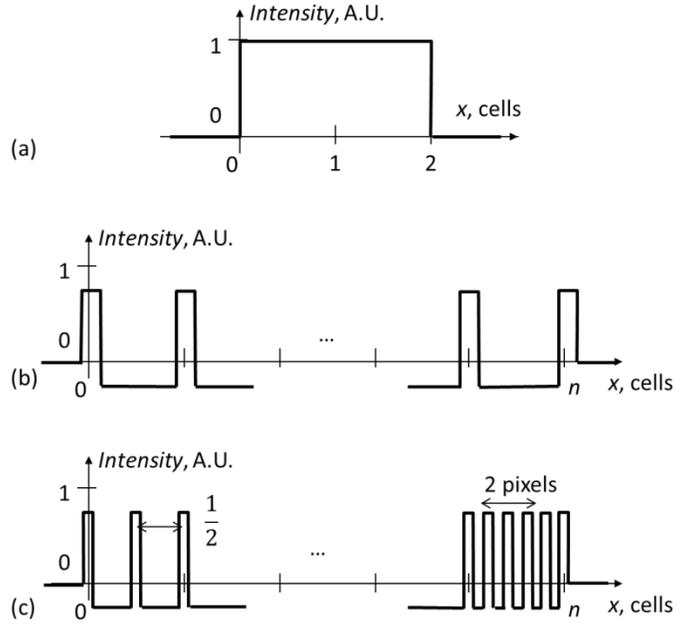

Fig. 11. Additional special "wavelets" to cover the whole spectrum. (a) scaling function, b) infinite plane, (c) chirp signal.

Similarly to the scaling function of the Haar wavelet, the rectangular pulse of the double width without the bias (see Fig. 11(a)) can be chosen as our scaling function,

$$\varphi_{1D} = \Pi(x/2) \qquad (29)$$

The normalization coefficient of the scaling function = ½.

The power spectra of the PB MV wavelets up to 16th order together with the scaling function are shown in Fig. 12. (Although, "period one" and "chirp" are not included, their place is clear.)

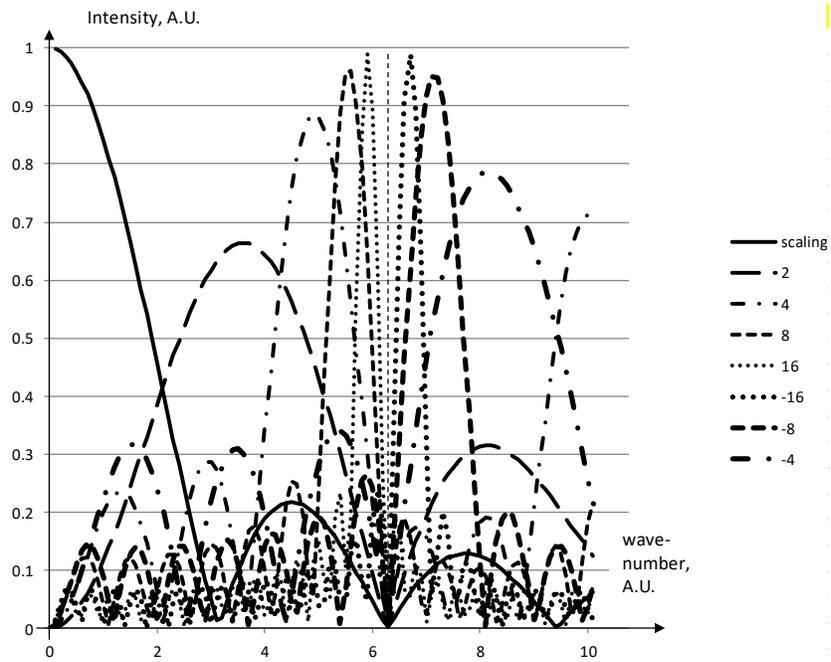

Fig. 12. Power spectra of PB wavelets up to 16th order together with scaling function.

### 3.2 2D wavelets

Consider the wavelets with the 2D parallax (FP) in a similar manner as 1D. In this case, we have the functions of two coordinates. 2D PB wavelets of the odd orders between -5 and +5 are shown in Fig. 13; the special cases between -2 and 2 in Fig. 14.

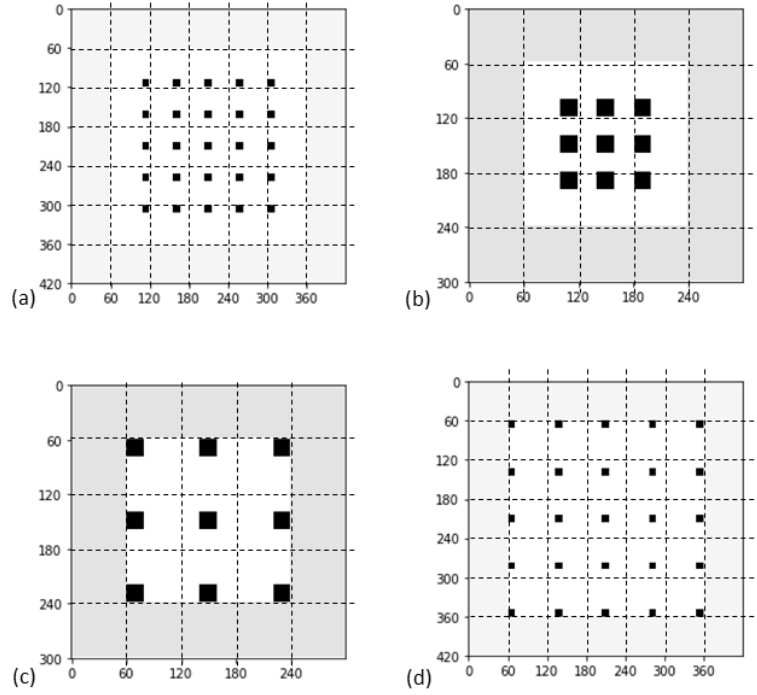

Fig. 13. 2D PB MV wavelets: (a) -5$^{th}$ plane, (b) -3$^{rd}$, (c) +3$^{rd}$, (d) +5$^{th}$. (Black color means maximum, white minimum, gray zero). The cells are shown by thin dashed lines.

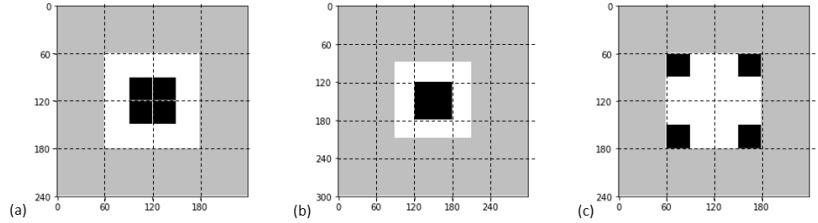

Fig. 14. 2D PB MV wavelets for -second, first, and +second depth planes in (a) – (c), resp. The cells are shown by thin dashed lines.

The width and period of pulses in each row/column of a 2D PB wavelet are the same as in the corresponding 1D case. Note the smaller bias and the changed normalization coefficient (as those for 1D Ricker and 2D Marr wavelets, both derived from the Gaussian). Namely, the bias of the 2D case is different because the wavelet is spread across $n \times n$ cells,

$$b_{2Dn} = \begin{cases} -\frac{1}{4}, n = 1 \\ -\frac{1}{n^2}, n > 1 \end{cases} \quad (30)$$

There are $n^2$ peaks of the height of $1-1/n^2$ (their total area = 1) plus the bias (height = $1/n^2$) across the area $n^2-1$ remaining for the bias. Therefore,

$$\|f\|^2 = \left(\left(1 - \frac{1}{n^2}\right) \cdot 1 + \frac{n^2-1}{n^2}\right)^2 = 4\left(\frac{n^2-1}{n^2}\right)^2 \quad (31)$$

and the normalization coefficient of 2D PB wavelets is

$$c_{2Dn} = \begin{cases} 1, & n = 1 \\ \frac{1}{2}\frac{n^2}{n^2-1}, & n > 1 \end{cases} \qquad (32)$$

The general formula for the 2D PB MV wavelets built as the normalized and biased product of the orthogonal 1D VPs is as follows,

$$\psi_{2Dn}(x,y) = c_{2Dn}\big(b_{2Dn} + P_n(x)P_n(y)\big) =$$

$$= c_{2Dn}\left(b_{2Dn} + \left(\sum_{i=0}^{n} \Pi(nx - ip_n - s_n)\right) \cdot \left(\sum_{j=0}^{n} \Pi(ny - jp_n - s_n)\right)\right) \quad (33)$$

where $c_{2Dn}$ is the normalization coefficient by Eq. (32), $b_{2Dn}$ is the bias by Eq. (30), VPs $P_n$ with their period and phase are given by Eqs. (3)-(4).

Based on the above formulas,

$$\psi_{2Dn}(x) = \frac{1}{2}\frac{n^2}{n^2-1}\left(-\frac{1}{n^2} + \sum_{i=0}^{n} \Pi\left(nx - i\frac{n+1}{n} - s_n\right) \cdot \sum_{j=0}^{n} \Pi\left(ny - j\frac{n+1}{n} - s_n\right)\right)(34)$$

The 2D scaling function is the product of two 1D scaling functions in each dimension,

$$\varphi_{2D} = \Pi(x/2) \ast \Pi(y/2) \qquad (35)$$

Its normalization coefficient $= ¼$ .

## 4. Numerical and visual experiments

In numerical experiments, the PB wavelets were used with the binary MV image [4] (two levels of brightness: the black and the white only) and with a gray-scale light-field image made by the independent author [17]. Calculations are verified and confirmed visually.

### 4.1. Direct transform

The first testing image is the tetrahedron. The base of the tetrahedron lies in the *xy*-plane and the camera (more exactly, an array of virtual cameras) is "in front of" the object, see Fig. 15(a). FP image of the tetrahedron is shown in Fig. 15(b) in pseudo colors: the white color in this and following illustrations of the tetrahedron means zero, the black color means one.

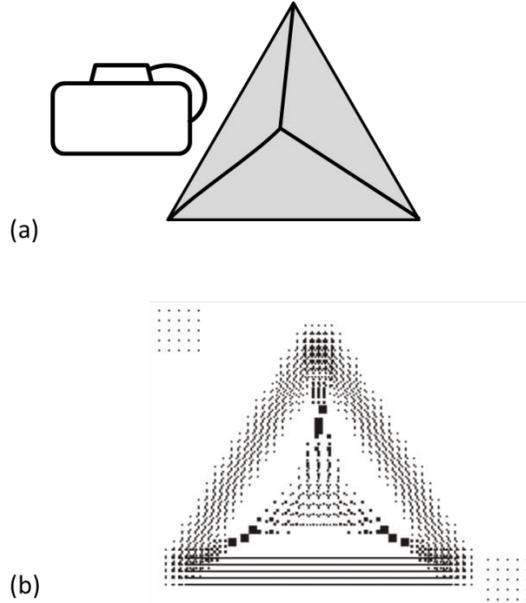

Fig. 15. Tetrahedron: (a) layout, (b) MV image.

For visual verification, the MV image was printed on a desk jet printer and displayed using two orthogonally crossed lenticular plates (15 lines per inch). Correspondingly, the size of the printed image cell was 1.69 mm. The visual appearance of the tetrahedron in 3D (technically, two photographs making up a stereo pair) is shown in Fig. 16.

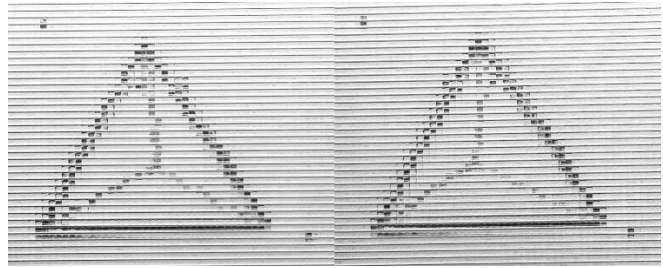

Fig. 16. Stereoscopic photograph of the original tetrahedron (crossed lenticular plates).

With certain practical skills, such a stereo pair can be seen with naked eyes, forming a 3D picture in our mind. Such common skills are not unknown [18], [19], [20]. The visual picture confirms that the vertex is closer to our eyes than the base.

The result of the direct transform is a 3D CWT array, which can be conveniently represented by planes (2D arrays). As compared to the conventional 1D wavelet transform of signals (along the timeline), each 2D array (a transformation across the image plane) is an analog of the line (at a single level) of the matrix of dilations/translations of that transform. A set of gray-scale images by planes is shown in Fig. 17; cross-sections of these 2D arrays along the thin lines in Fig. 17 are shown in Fig. 18.

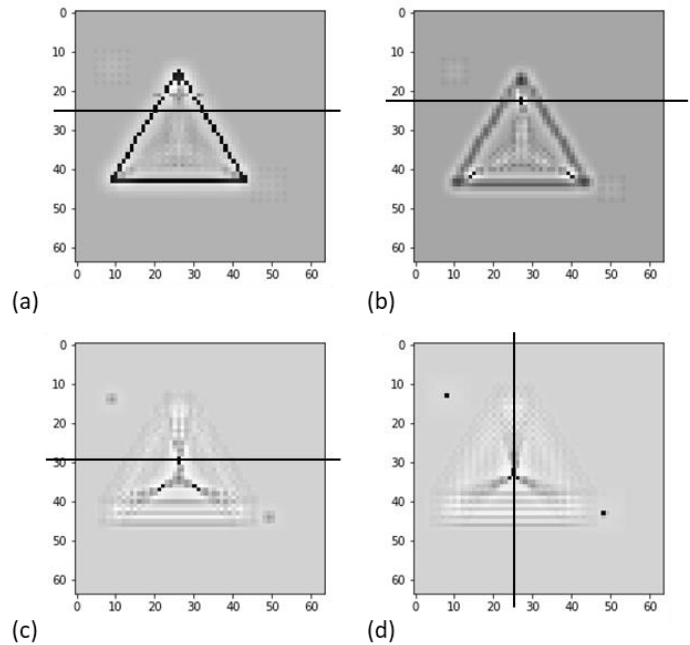

Fig. 17. Arrays of CWT coefficients for tetrahedron (a)-(d): $d$ = -5, -3, 3, 5. (black = maximum, white = minimum, gray = near zero).

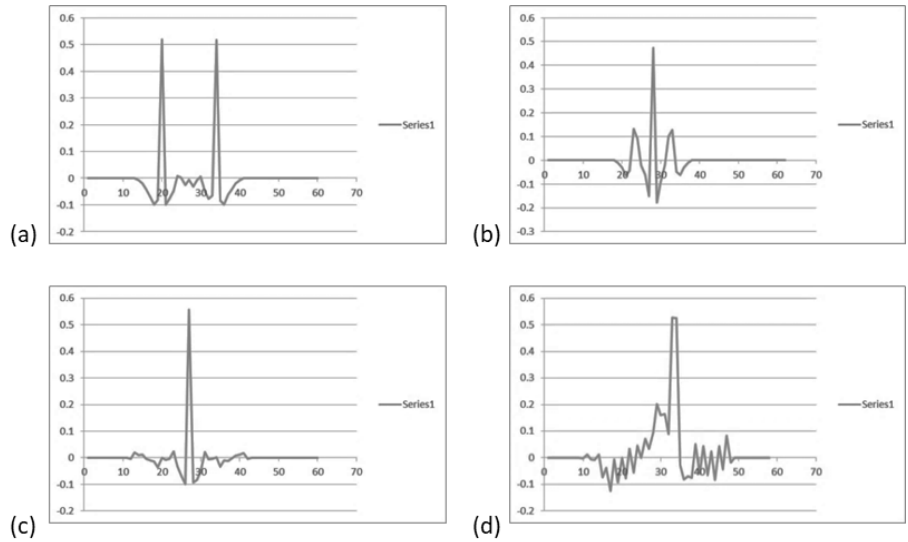

Fig. 18. Cross sections of CWT arrays for tetrahedron, (a)-(d): $d$ = -5, -3, 3, 5.

Another testing image is the image of books on a desk, which is available for free [17]; see Fig. 19.

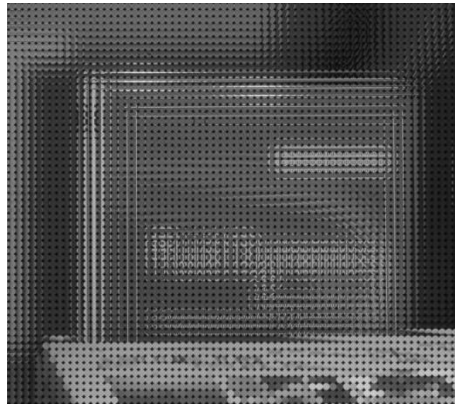

Fig. 19. Magnified part of original light-field image of books. Credits to T. Georgiev <www.tgeorgiev.net>.

The results of the direct wavelet transform of this light-field, i.e. 2D arrays of the wavelet coefficients at two depth planes are shown in Fig. 20. In this case, the colors (levels of gray) correspond to the original image.

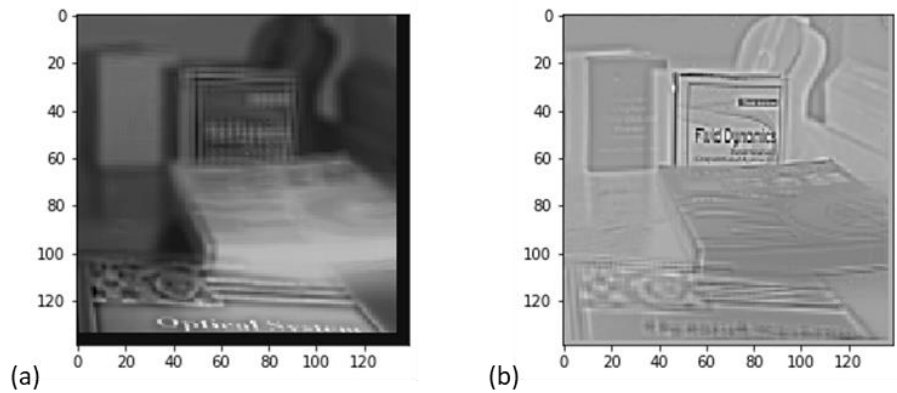

Fig. 20. 2D arrays of CWT coefficients for -6th and +3rd planes. Pseudo colors as in Fig. 17.

The titles of the books in Fig. 20 ("Optical System" and "Fluid Dynamics") can be clearly read in the corresponding planes, while the content of other planes is blurred. This shows the selectivity of the proposed wavelet transform by depth.

### 4.2 Direct and inverse transforms

The result of the direct transform can be used as the input of the inverse wavelet transform. We use the same wavelet functions in both direct and inverse transforms, although this might not be the most general option. The inverse transforms of all planes can be merged (in a simple case like this, it is added) into a single computer-generated MV image. Thus, we may restore a whole 3D object from the arrays of coefficients by planes. The restored object is shown in Fig. 21. Note that with the scaling function, the positive intensity range [0, 1] seems to be restored correctly.

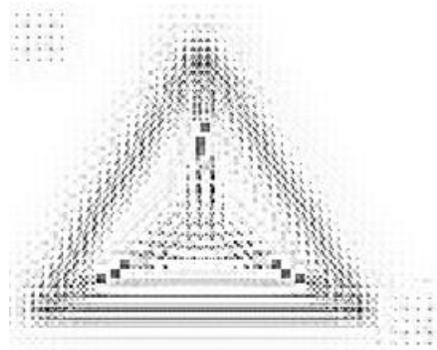

Fig. 21. Restored MV image of the tetrahedron.

The restored MV image looks similar to the original, as expected; compare the MV images in Figs. 15(b) and 21. The computer-generated restored image was displayed in 3D; the result is shown in Fig. 22.

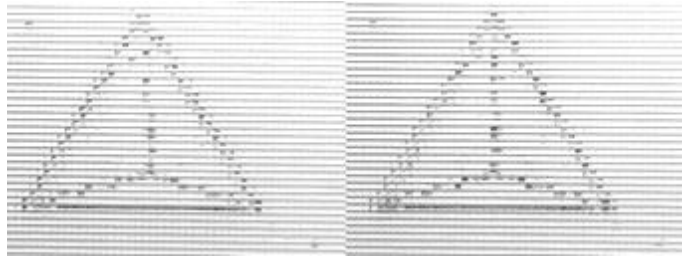

Fig. 22. Stereoscopic photograph of the restored tetrahedron.

Compare the photographed image of the original tetrahedron in Fig. 16 and its restored counterpart in Fig. 22. The visual observation confirms that the restored 3D object is identical to the original.

For the image of books, the restored image reproduces the structure of the original light-field correctly as shown in Fig. 23(a), although we did not take into account an essential vingetting of each cell of the original light-field image. Anyway, to mimic the vingetting, the rectangular array of circular holes was applied after the wavelet processing, see Fig. 23(b). Compare the corresponding parts of the images in Figs. 19 and 23(b).

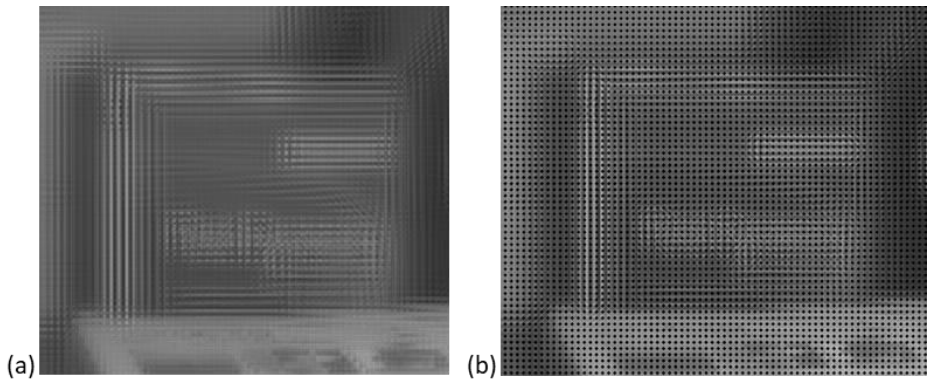

Fig. 23. (a) Restored light-field image after inverse wavelet transform. (b) Restored image with the mask. (Magnified as Fig. 19.)

### 4.3 Direct transform, modification, and inverse transform

Furthermore, as we obtained the 3D array of CWT coefficients, we may modify it somehow and it again as the input of the inverse transform. In this way, we can construct other 3D objects from the modified coefficients. We made two modifications: a) the reversed depth and b) the dimensions of the parallax (2D parallax → 1D parallax).

In the first modification, the sign of the depth of each 2D CWT array was reversed and the inverse wavelet transform was applied. The MV image of the tetrahedron with the reversed depth is shown in Fig. 24. Note that such a modification looks good for a wireframe 3D object, but could be improper for a textured object because in that case, it may cause a so-called "pseudoscopic effect".

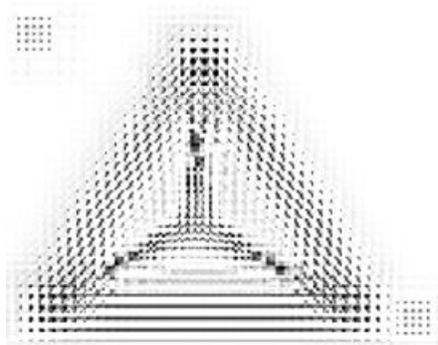

Fig. 24. MV image of the tetrahedron with reversed depth.

To find differences in the image plane, compare the MV image shown in Fig. 24 with Fig. 16(b). The stereoscopic photograph of the displayed tetrahedron with the reversed depth is shown in Fig. 25 (as in the previous case, the 3D object was displayed using the crossed lenticular plates). In the 3D object with the reversed depth, the visual position of the vertex is behind the base (farther from an observer), as expected. Compare the photographed image with its original counterpart in Fig. 16.

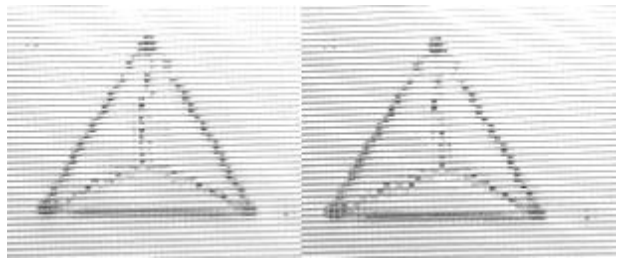

Fig. 25. Stereoscopic photograph of the tetrahedron with reversed depth.

In the second modification, the dimensions of the parallax were changed, i.e., the 3D CWT array (some planes are shown in Fig. 17) was used in the inverse wavelet transform with 1D wavelets instead of 2D wavelets in Fig. 21. The resulting MV image with the modified dimensions of the parallax (2D → 1D) is shown in Fig. 26.

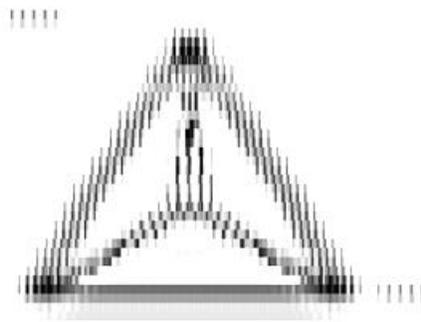

Fig. 26. MV image of the tetrahedron with the modified dimension of parallax (FP → HPO).

To display this MV image with HPO, we used one lenticular plate. The stereoscopic photograph of the displayed HPO tetrahedron is shown in Fig. 27.

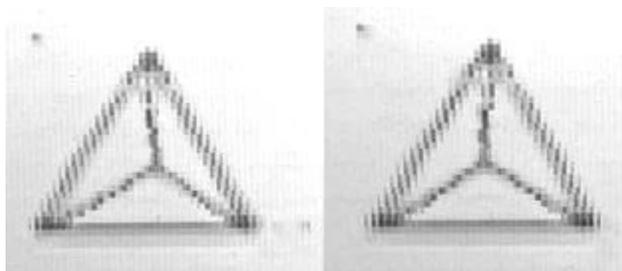

Fig. 27. Photograph of the tetrahedron with the modified dimension of parallax (2D → 1D).

It can be clearly seen that that the structure of the object remains the same after the modification of the parallax dimensions. The position of the vertex in front of the base is confirmed. Compare the photograph of the modified object in Fig. 27 with its original counterpart in Fig. 16.

## 5. Discussion

We performed the continuous wavelet transform of MV images with 1D and 2D parallax (displayed using the vertical and orthogonal lenticular plates). This way, we model a small area of a whole screen of ASD. Due to the optical similarity of the lens camera and the pinhole camera, the results are also valid for ASD with the barrier plate.

The direct and inverse wavelet transforms using the proposed wavelets (as well as [13], [14]) are not a universal instrument to analyze/synthesize arbitrary images. We should emphasize that they only work well with the MV images. Applying these wavelets to other sorts of images might be meaningless as, e.g., the fresnelets [6], [7] are only useful with Fresnel holograms.

A couple of examples of the entire cycle (direct and inverse transform) in ASD could be as follows. To find if a 3D object approaches another 3D object, we may analyze the depth of the underlying surface in real-time by the wavelet analysis, and when the depth of the surface (the result of analysis) would be close to the depth of the surface of another object, we may, e.g., re-synthesize one surface (recolor it, make it blink, etc.), and show the synthesized object in 3D. This way, we may draw additional attention of the operator to an approaching object. In another example, we may prevent one 3D object from "penetrating" into another 3D object [21]. When the first object is behind the second object in the 3D space (by the depth analysis of the wavelet coefficients), we may either hide it (i.e., refrain from drawing) or transform somehow (e.g., re-synthesize it with a reduced contrast).

Some practical requirements of this technique are as follows. The elementary view images of the MV image in the image plane should be rectified; at least because VPs in non-rectified MV images were not observed. The square cell should comprise an integer number of pixels ($n^2$ pixels). MV image should comprise whole cells; at least, the whole cell must lie in the left upper corner. However, the latter is not a fundamental restriction, just a limitation of the current version of the computer program.

## 6. Conclusion

We built the multiview wavelets based on the voxel patterns and describe the wavelets in the spatial and spectral domains. The direct and inverse wavelet transforms were demonstrated in binary/gray-scale images with the 1D/2D parallax (horizontal parallax only and full parallax, resp.). We also provided examples of other modifications in 3D using the proposed multiview wavelets: the reversed depth and the changed dimension of the parallax. The correctness of the restored image is verified in each case visually. The results can be used in the multiview/light-field image processing in general, as well as in the depth analysis and the 3D interaction in particular.